\pdfoutput=1

\documentclass[10pt,letterpaper]{article}

\usepackage{cogsci}
\usepackage{pslatex}
\usepackage{apacite}
\usepackage{amsmath}

\usepackage{graphicx}
\usepackage{natbib}

\newcommand{\comment}[1]{}

\title{Adapting Deep Network Features to Capture Psychological Representations}

\author{
{\large \bf Joshua C. Peterson (jpeterson@berkeley.edu)} \\
{\large \bf Joshua T. Abbott (joshua.abbott@berkeley.edu)}\\
{\large \bf Thomas L. Griffiths (thomas\_griffiths@berkeley.edu)} \\
  Department of Psychology, University of California, Berkeley, CA 94720 USA}

\begin{document}

\maketitle

\begin{abstract}
Deep neural networks have become increasingly successful at solving classic perception problems such as object recognition, semantic segmentation, and scene understanding, often reaching or surpassing human-level accuracy. This success is due in part to the ability of DNNs to learn useful representations of high-dimensional inputs, a problem that humans must also solve. We examine the relationship between the representations learned by these networks and human psychological representations recovered from similarity judgments. We find that deep features learned in service of object classification account for a significant amount of the variance in human similarity judgments for a set of animal images. However, these features do not capture some qualitative distinctions that are a key part of human representations. To remedy this, we develop a method for adapting deep features to align with human similarity judgments, resulting in image representations that can potentially be used to extend the scope of psychological experiments.

\end{abstract}

\section{Introduction}
The resurgence of neural networks in the form of \textit{deep learning} has continued to dominate object recognition benchmarks in the field of computer vision, often attaining near or above human-level accuracy for a variety of perceptual tasks, most notably through recent advances in classifying thousands of objects within natural images \citep*{krizhevsky2012imagenet,he2015delving}. Part of the success of these models is due to their ability to learn effective feature representations of high-dimensional inputs (e.g., complex color images); a challenge that human perception must also confront \citep*{austerweil_nonparametric_2013}. As a result, cognitive scientists have started to explore how the representations learned by these networks can be used in models of human behavior for perceptual tasks such as predicting the memorability of objects in images \citep*{ICCV15_ObjectMemorability} and predicting judgments of category typicality  \citep*{lake_deep_2015}.

While deep learning models continue to mimic a growing list of human-like abilities, a number of core questions remain unanswered about the relevance of these models to actual human cognition and perception. For instance, features of the input learned using these networks excel in predicting certain human judgments, but how are these feature representations related to human psychological representations? At first glance, it would seem that the ability of these representations to predict typicality judgments and stimulus memorability would constitute robust evidence of their relevance to people, however recent work has shown that neural networks that classify images can be systematically deceived by imperceptible image transformations \citep*{szegedy2013intriguing}, casting doubt on their similarity to humans.

Understanding the relationship between the representations found by deep learning and those of humans is an important question in cognitive science, and could potentially benefit artificial intelligence. However, independent of this question, simply having a good approximation to how people represent images would allow cognitive scientists to test psychological theories using complex, realistic stimuli. Indeed, tasks such as creating stimulus sets that uniformly span psychological space are far from trivial.

In this paper, we address this question directly by examining how well features extracted from state-of-the-art deep neural networks predict human similarity judgments. An initial evaluation shows that these features account for a significant amount of variance in human judgments, but fail to capture qualitative distinctions that are key to human representations. We then develop a method for adapting deep network features to better predict human similarity judgments, and show that this approach can reproduce those qualitative distinctions. These results suggest that while raw features produced by deep learning may not be suitable for use in modeling cognition, they can be modified to bring them into close alignment with human representations.

\section{Deep Representations}
In general, deep neural networks (DNN) are neural networks that have depth in terms of their number of hidden layers between input and output \citep*{bengio2009learning}. In the past few years, training such networks to understand aspects of large, complex data sets has led to a number of advances in vision and language applications \citep*{lecun2015deep}.

In computer vision, the majority of this progress has been driven by a particular DNN called a convolutional neural network (CNN) \citep*{lecun1989backpropagation}. CNNs get their name from the use of convolutional layers, which learn a set of image filters that produce feature maps of spatially-organized inputs like images. This allows for a drastic decrease in the number of parameters the network must learn, which would otherwise explode exponentially in a fully connected network with high-dimensional inputs. The typical CNN architecture includes a series of hidden convolutional layers, followed by a smaller number of fully connected layers, and finally a layer that generates the final output or classification. While CNNs were initially developed over two decades ago, they came to mainstream popularity in 2012 when a 7-layer architecture named AlexNet \citep*{krizhevsky2012imagenet} won the ImageNet Large Scale Visual Recognition Challenge (ILSVRC), reducing the previous winner's error rate by an uncommonly large margin. Since then, a deeper CNN has won the contest every year, currently dominated by Microsoft's 150-layer network which obtained a best-of-top-5 error rate of 4.94\%, surpassing the accuracy of non-expert humans at 5.1\% \citep*{he2015delving}.

Interestingly, CNNs produce much more than just their outputs (e.g., a category label for an image); they can also return feature representations at each layer of the network. The ``deep representations'' learned by these networks have proven useful in predicting human behavior. \cite*{ICCV15_ObjectMemorability} used representations extracted from the last fully-connected layer of a CNN to predict the intrinsic memorability of objects. That is, the objects that humans are jointly likely to remember or forget in a large complex natural scene database. The correlation between estimates of memorability and the original memorability scores for each object matched human consistency (i.e. the correlation between memorability scores of random splits of the full sample of subjects). Similarly, \cite*{lake_deep_2015} were able to reliably predict human typicality ratings of eight object categories using the same network and features, and called for cognitive scientists to pay attention to deep learning since categorization is a foundational problem in the field.

Deep representations are also beginning to interest the neuroscience community. For example, CNN activations have been used to predict monkey IT cortex activity \citep*{yamins2014performance}, as well as both low- and high-level activity in human visual areas \citep*{agrawal_pixels_2014}. Delving deeper, \cite*{khaligh-razavi_deep_2014} found that a CNN best explained IT cortex representations out of a set of 37 well-known models from both the computer vision and neuroscience fields, although no model completely explained all of the variance, unsupervised models being the worst of all of them.

Although CNN representations currently do the best job of predicting neural activity as measured by Blood Oxygenation Level Dependent (BOLD) response, this does not guarantee that we can explain psychological representations as a result. In fact, \cite*{mur_human_2013} was partly successful in predicting human similarity judgments (a classic index of psychological representations) from IT cortex representations. However, the key categorical distinctions in the human representations were not well predicted: human IT cortex representations were more similar to monkey IT cortex representations than they were to human psychological representations. In the remainder of the paper, we use a similar approach to evaluate how well deep network features align with human psychological representations, and to explore how the correspondence between the two can be increased.

\section{Evaluating Representations}
Our first step is to evaluate the potential correspondence between deep network features and psychological representations. Unlike neural representations, psychological representations cannot be measured directly. However, both spatial and hierarchical psychological representations for $N$ objects can be recovered given an $N \times N$ matrix of similarity judgments using methods such as multidimensional scaling and hierarchical clustering \citep*{shepard1980multidimensional}. We thus reduce the problem to one of capturing human similarity judgments, subjecting both human judgments and model predictions to these different methods of extracting representations. We approach this problem by taking the inner-product of the deep feature representations of each pair of images (a measure of similarity between two vectors). We then compute the correlation between these pairwise vector similarities and human similarity judgments for the same stimulus pairs, which gives us a measure of the correspondence we want to evaluate.\\

\begin{figure*}[!ht]
\begin{center}
\includegraphics[width=\linewidth,keepaspectratio]{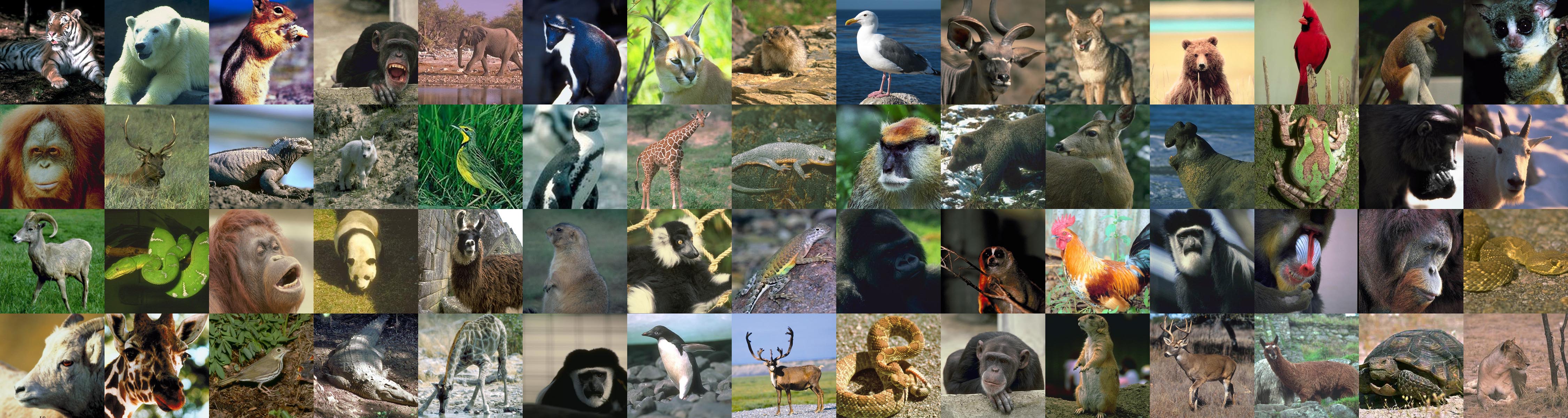}
\end{center}
\vspace{-5px}
\caption{Samples from the set of $120$ animal photographs.}
\label{stimulus-samples}
\end{figure*}

\noindent \textbf{Stimuli.} Our stimulus set consisted of $120$ color photographs of animals (sample images are shown in Figure \ref{stimulus-samples}). Images were cropped to $300\times300$ pixels, resulting in close-ups of either the animal's face or body. The set was constructed to include both inter- and intraspecies variation.\\
%
%

\noindent \textbf{Behavioral Experiment.} We collected pairwise similarity ratings for our animal stimulus set through Amazon Mechanical Turk. Participants were instructed to rate the similarity of four pairs of animal images on a scale from $0$ (not similar at all) to $10$ (very similar). We paid workers $\$0.02$ per set of four comparisons. Before each task, eight examples were shown to help prevent bias in early judgments. Amazon workers could repeat the task with new animal pairs as many times as they wanted. There were $7,140$ possible image comparisons, each of which was rated by 10 unique participants, for a total of $71,400$ ratings from $209$ different participants. The result was a $120\times120$ similarity matrix after averaging over judgments.\\

\noindent \textbf{Feature Extraction.} We extracted features for each image in our data set using three different popular \textit{off-the-shelf} CNNs of varying complexity that were pretrained in Caffe \citep*{jia2014caffe}. Specifically, we used CaffeNet (based on original AlexNet), VGG16 \citep*{simonyan2014very}, and GoogLeNet \citep*{szegedy2014going}, the layer depths of which were $7$, $16$, and $22$ respectively. GoogLeNet and VGG16 achieve roughly half the error rates of AlexNet. Each network had already been trained to classify 1000 object categories from previous ILSVRC competitions. A feedforward pass of each flattened image vector into each network yields feature responses at each layer. For our analysis, we extracted the last layer of each network before the classification layer. For CaffeNet and VGG16, this is a 4096-dimensional fully-connected layer, while the last layer in GoogleNet is a 1000-dimensional average pooling layer. Lastly, we also extracted Histograms of Oriented Gradients (HOG) and Scale-Invariant Feature Transform (SIFT) representations for comparison since such features represent the generic representations of choice for tasks in computer vision prior to the popularity of deep learning. \\

\begin{table}[!t]
\caption{Correlations between human and deep similarities.}
\begin{center}
\label{vanilla-results}
\begin{tabular}{lcccc}
          & CaffeNet & Google & VGG & HOG+SIFT \\
\hline
$R^{2}$   & .32 & .35 & .43 & .008  \\
\end{tabular}
\end{center}
\end{table}

\subsection{Results} Table \ref{vanilla-results} gives performance ($R^{2}$) for each model. Raw representations from all three networks show medium to high correlations with the human data. In general, deeper networks with better ImageNet classification accuracy like GoogLeNet and VGG16 did better than CaffeNet, which is considerbly more shallow. The HOG+SIFT baseline did surprisingly poorly, explaining very little variance as compared to the deep representations, suggesting that while these features are useful for many computer vision tasks, they differ in large part from the representations humans employ when judging animal similarity.

\begin{figure*}[!ht]
\begin{center}
\includegraphics[width=\linewidth,keepaspectratio]{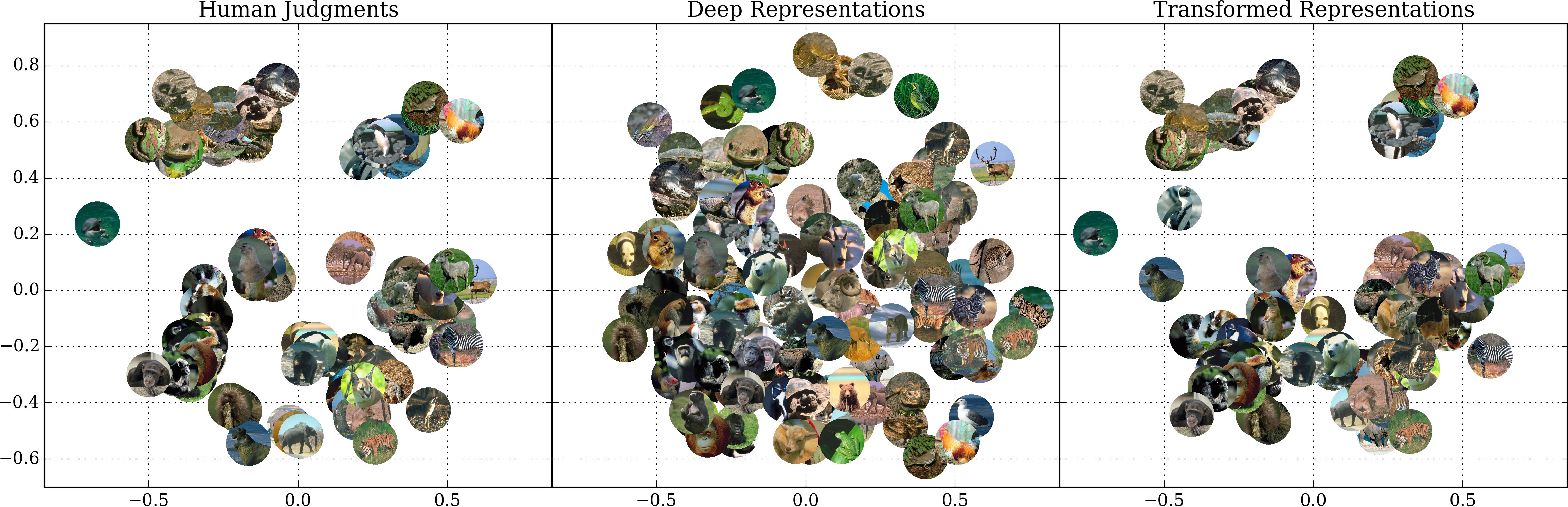}
\end{center}
\vspace{-5px}
\caption{Multidimensional scaling solutions for similarity matrices obtained from human judgements (left), non-transformed deep representations (center), and transformed deep representations (right).}
\label{mds-plots}
\end{figure*}

\begin{figure*}[!ht]
\begin{center}
\includegraphics[width=\linewidth,keepaspectratio]{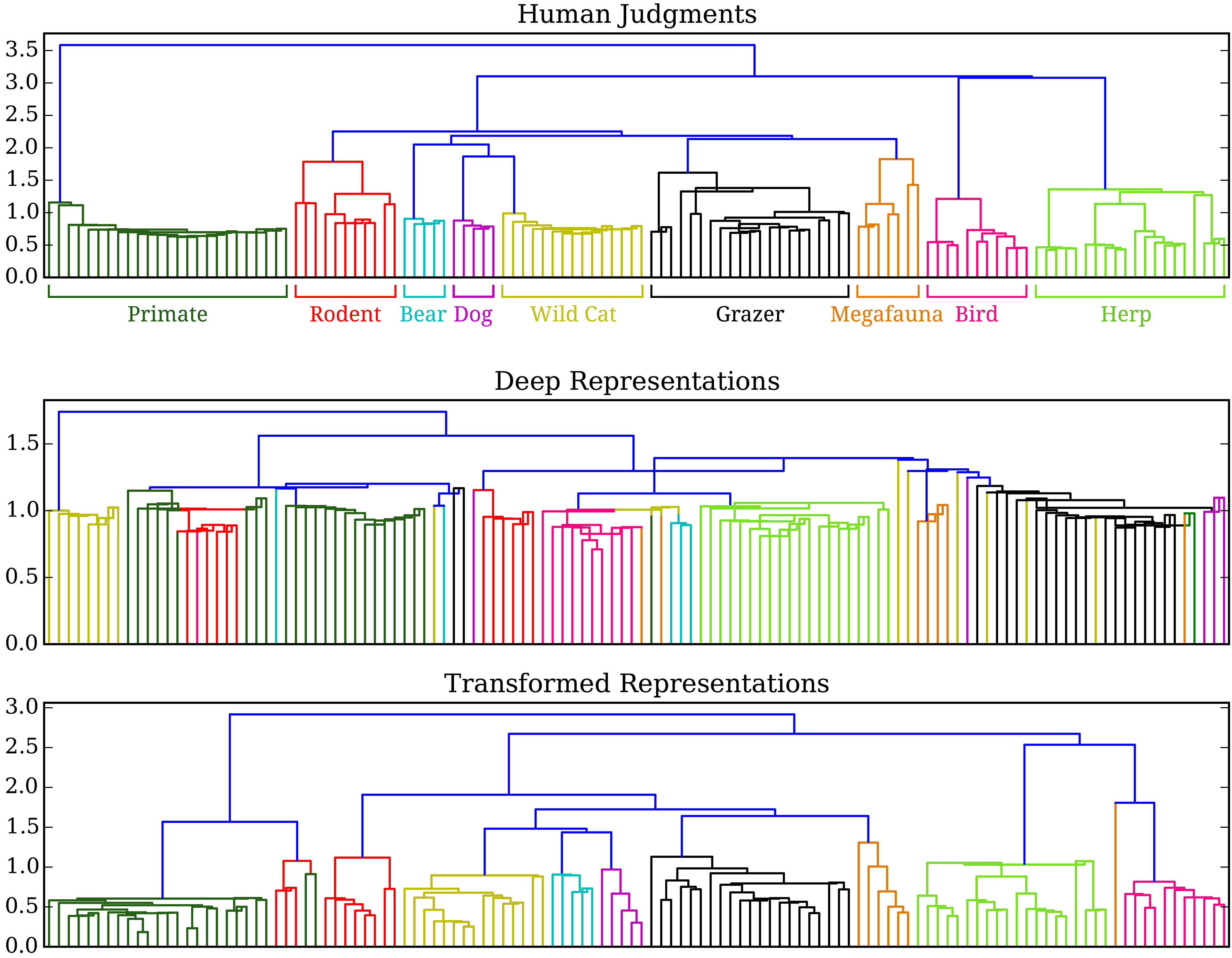}
\end{center}
\vspace{-5px}
\caption{Hierarchical clustering of human judgements (top), deep representations (middle), and transformed representations (bottom). Human judgments resulted in nine interpretable clusters, grouped by color and semantic category label in the top panel. The leaves of the deep and transformed representation clusterings are color-coded relative to the human judgments.}
\label{dend-plots}
\end{figure*}

Although the VGG representation explained a fair amount of variance, further analyses revealed that the most crucial structural aspects of the human representations were not preserved. The first and second panels of Figure \ref{mds-plots} show multidimensional scaling (MDS) solutions for the original human data and the predictions from the unaltered deep representations. While the structure of the MDS solutions for the predicted judgments looks reasonable (e.g., zebras are next to other zebras), major categorical divisions are not preserved. Hierarchical clusterings of the actual and predicted human judgments (the first and second panels of Figure \ref{dend-plots}) show a similar pattern of results: human judgments exhibit several major categorical divisions, whereas much of this structure is lost in the predicted data.

\section{Adapting Representations}
After quantifying the discrepancy between deep and human representations, we can attempt to bring them into closer alignment. First, consider that the final hidden layer feature representation in a neural network can be thought of as the input to a final linear classification layer, such that the problem solved by the final weight matrix is a linear transformation (which is then often scaled by a softmax function to covert to class probabilities). This can be thought of as a rescaling of the final stimulus representation to solve the categorization problem. This suggests that we should not think about the features extracted by the network as a static representation, but as the ingredients for a transformation that solves a problem. Thinking in these terms, we show that we can easily solve for a linear transformation that better captures human similarity judgments. \\

\noindent \textbf{Similarity Model.} Any similarity matrix ${\bf S}$ can be decomposed into the matrix product of a feature-by-object matrix ${\bf F}$, its transpose, and a diagonal weight matrix ${\bf W}$,
\begin{equation}
    {\bf S} = {\bf F}{\bf W} {\bf F}^T
\end{equation}
This formulation is similar to that employed by additive clustering models \citep*{shepard1979additive}, wherein {\bf F} represents a binary feature identity matrix (and is similar to Tversky's (1977) model of similarity). When used with continuous features, this approach is akin to factor analysis. Given an existing feature-by-object matrix {\bf F}, the diagonal of ${\bf W}$ can be solved for using linear regression where the predictors for each similarity $s_{ij}$ are the product of the values of each feature for the objects $i$ and $j$. When ${\bf W}$ is the identity matrix, this reduces to the model evaluated in the previous section.
\begin{equation}
  \label{eq:u}
  \begin{gathered}
\displaystyle s_{ij} = \sum_{i=1}^{N_{f}} w_{k}f_{ik}f_{jk} .
 \end{gathered}
 \end{equation}
The result is a convex optimization problem that can be solved straightforwardly, allowing us to find a transformation of the deep features with a closer correspondence to human similarity judgments. \\

\noindent \textbf{Analysis.} With such a large number of predictors, regularization is critical to avoid overfitting. We used ridge regression ($L2$ regularization) and performed grid search on cross-validated generalization performance to find the best regularization parameter. We predicted only the upper triangle of the similarity matrix since the matrix is symmetric. Each model was evaluated via its generalization performance in 6-fold cross-validation. We did this for the feature vectors extracted at each layer of the network.

As an additional control against overfitting, we compared model performance with several baselines. In Baseline 1, we shuffled the rows of the feature matrix such that the feature representation of one image was replaced with that of a different randomly chosen image. In Baseline 2, the columns of the feature matrix were randomly permuted for each row separately. Lastly, Baseline 3 simply combined the shuffling schemes from the first two baselines. In all three cases, the randomized feature matrices were subjected to the same set of analyses as the true features, allowing us to check for spurious correlations. \\

\subsection{Results} Table \ref{adjust-results} shows performance for each network using our adjustment of the\comment{CNN} representations. The $R^{2}$ values reported are the average values across all six folds of the crossvalidation. All five models performed considerably well, each showing improvement over the original non-weighted models. Most notably, VGG16 performed best, accounting for 84\% of the variance. Training using the estimated regularization parameter on the entire dataset yielded an $R^{2}$ of 91\%. In contrast, all three baseline models explained essentially no variance ($R^{2}<0.01$), suggesting that our results were not spurious correlations resulting from our large sets of predictors. 

Crucially, the MDS solution for the improved predictions is almost identical to the original human spatial representation. The same improvements were found in hierarchical clusterings of actual and predicted similarity matrices (1st and 3rd panels of Figure \ref{dend-plots}), this time largely in the form of top-level parent nodes. \\

\begin{table}
\begin{center}
\caption{Model performance using adjusted CNN features.}
\label{adjust-results}
\vskip 0.12in
\begin{tabular}{lcccc}
      & CaffeNet & Google & VGG & SIFT \\
\hline
$R^{2}$ & .69  & .72 & .84 & .09  \\
\end{tabular}
\end{center}
\end{table}

\begin{figure}
\begin{center}
\includegraphics[width=\linewidth,keepaspectratio]{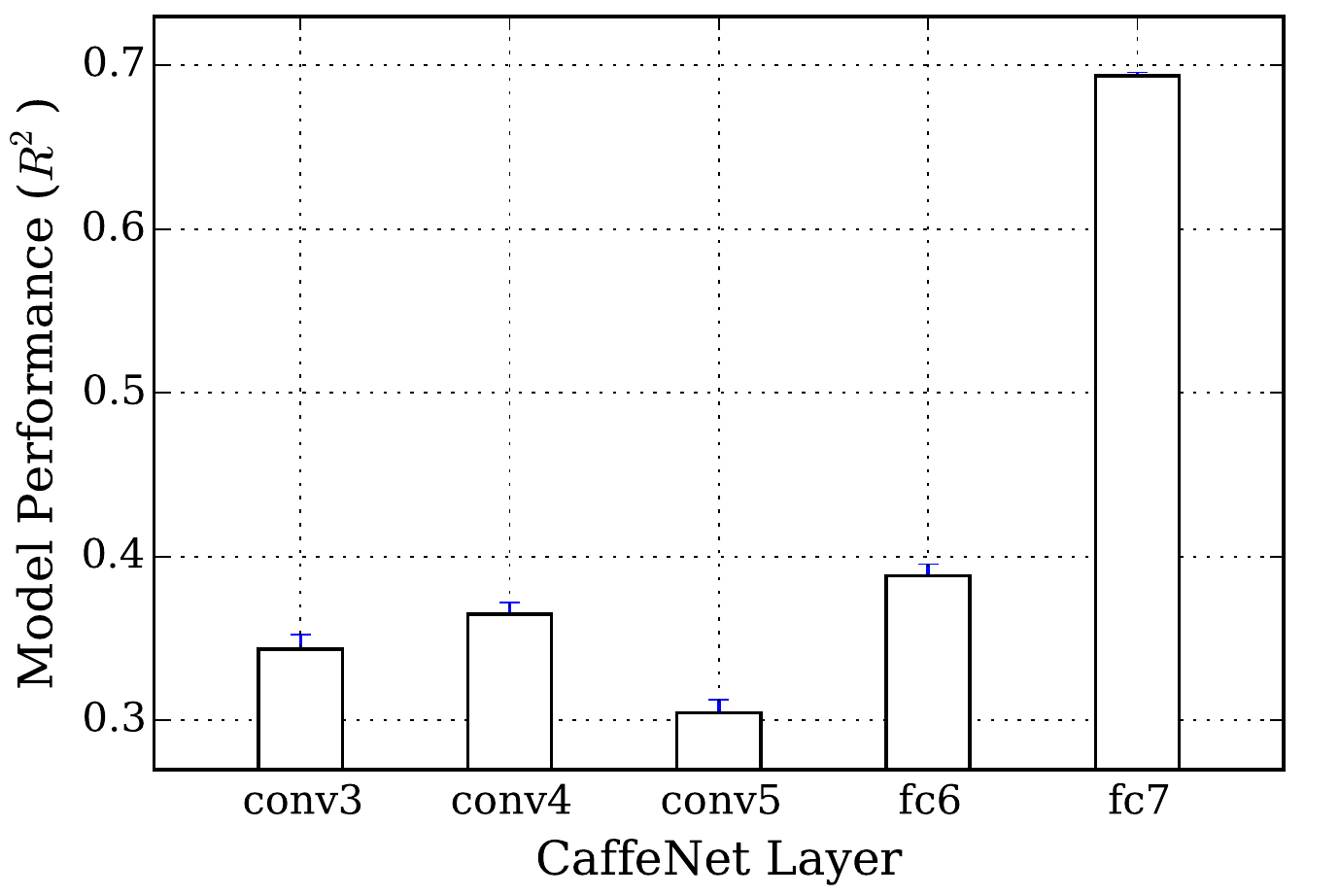}
\end{center}
\vspace{-10px}
\caption{Model performance as a function of feature layer depth in CaffeNet.}
\label{model-performance}
\end{figure}

\noindent\textbf{Feature Analysis.} While higher layers in CNNs tend to produce the most generic high-level features for domain transfer across image applications, the choice of feature depth is ultimately dependent on the task \citep*{sainath2013learning}. This implies that layer responses at different depths may explain different types of human similarity judgments (e.g. tasks that involve comparing visual features versus conceptual information). We examined our model's performance in predicting similarity judgments as a function of feature depth using CaffeNet, given its more straightforward architecture and manageably-sized layers. Specifically, we compared performance across the last three convolutional layers and the last two fully-connected layers. The results are shown in Figure \ref{model-performance}. Performance does appear to correspond strongly to layer depth, although fully connected layers perform much better than convolutional layers, suggesting that human similarity judgments may not be explained well from simpler image features.\\

\noindent\textbf{Reweighted Classification.} We investigated the effect of our fine-tuned representations on a separate animal classifier, using a new animal data set consisting of $1,740$ images from $19$ animal classes (\textit{bear, cougar, cow, coyote, deer, elephant, giraffe, goat, gorilla, horse, kangaroo, leopard, lion, panda, penguin, sheep, skunk, tiger, zebra}) \citep*{afkham2008joint}. We used multinomial logistic regression with 6-fold cross-validation to classify animals using fine-tuned representations as predictors. We fine-tuned these representations by pairwise multiplying the original VGG16 representations with the square-root of the weights obtained through prediction of the human similarity data. However, because some of the weights of the original solution are negative, we used Elastic Net regression to solve for weights constrained to be positive. We ran the same model using the original unaltered VGG16 representation to serve as baseline performance. The original model performed very well (average $R^2=.94$), whereas the fine-tuned model performed consistently worse ($R^2=.89$).
\section{Discussion}
This analysis constitutes the first formal comparison of deep representations to human psychological representations. Initial results using currently high-performing convolutional neural networks show that the two representations were moderately correlated, but diverge in terms of crucial structural characteristics, a problem exhibited by similar experiments using neural representations as opposed to deep features \citep*{mur_human_2013}.

Our method of overcoming this problem, by a parsimonious adjustment of the feature representation inspired by a classic model of similarity, appears to have been largely successful. Indeed, the human representations were almost completely reconstructed by our adjusted CNN features. Using features extracted from deep convolutional neural networks provides an opportunity to estimate psychological representations from real, raw sensory inputs (e.g. pixels). However, one potential limitation of this work is the generalizability of the transformation acquired to broader stimulus contexts. Testing this question will require replication and transfer across several domains. To the extent that this can be established, we envision our method as a standard tool for studying cognitive science using natural stimulus sets, on par with modern artificial intelligence.

Beyond this, we see potential for such an interface between cognitive science and artificial intelligence to be exploited for the benefit of each. While our attempt to improve a common categorization objective in computer vision (i.e. one-versus-all classification) using human-tuned representations was not successful, it does raise interesting distinctions between the computational problems solved by humans and CNNs. After all, the full breadth of human categorization behavior exhibits complex patterns such as overlapping class assignments, which are not likely to be well-represented when the learning objective is defined through images and objects characterized by a single label. Further, one might ask if poor categorization performance of the one-versus-all kind is the price paid for a more flexible system of categorization with respect to a set of complex objects that can be partitioned using several ``good'' configurations, depending on the context and task at hand. Given this possibility, one should be careful not to equate CNN classification performance with human categorization abilities in general.
\nocite{afkham2008joint}
\nocite{agrawal_pixels_2014}
\nocite{austerweil_nonparametric_2013}
\nocite{he2015delving}
\nocite{jia2014caffe}
\nocite{khaligh-razavi_deep_2014}
\nocite{krizhevsky2012imagenet}
\nocite{lake_deep_2015}
\nocite{lecun1989backpropagation}
\nocite{mur_human_2013}
\nocite{ICCV15_ObjectMemorability}
\nocite{sainath2013learning}
\nocite{shepard1979additive}
\nocite{simonyan2014very}
\nocite{szegedy2013intriguing}
\nocite{szegedy2014going}
\nocite{yamins2014performance}
\nocite{bengio2009learning}
\nocite{lecun2015deep}

\vspace{2mm}
\begin{small}
\noindent{\bf Acknowledgments.} This work was supported by grant number FA9550-13-1-0170 from the Air Force Office of Scientific Research. We thank Alex Huth for help with image selection.
\end{small}

\bibliographystyle{apacite}


\setlength{\bibleftmargin}{.125in}
\setlength{\bibindent}{-\bibleftmargin}

\bibliography{CogSci_Template}

\end{document}